\documentclass[conference]{IEEEtran}
\usepackage{cite}
\usepackage{amsmath,amssymb,amsfonts}
\usepackage{algorithmic}
\usepackage{graphicx}
\usepackage{textcomp}
\usepackage{xcolor}
\usepackage{booktabs}
\def\BibTeX{{\rm B\kern-.05em{\sc i\kern-.025em b}\kern-.08em
    T\kern-.1667em\lower.7ex\hbox{E}\kern-.125emX}}

\usepackage{eso-pic}
\newcommand\AtPageUpperMyright[1]{\AtPageUpperLeft{
 \put(\LenToUnit{0.5\paperwidth},\LenToUnit{-1cm}){
     \parbox{0.5\textwidth}{\raggedleft\fontsize{9}{11}\selectfont #1}}
 }}
\newcommand{\conf}[1]{
\AddToShipoutPictureBG*{
\AtPageUpperMyright{#1}
}
}

\makeatletter
\def\ps@IEEEtitlepagestyle{
  \def\@oddfoot{\mycopyrightnotice}
  \def\@evenfoot{}
}
\def\mycopyrightnotice{
  {\footnotesize \hfill 978-1-7281-5842-6/19/\$31.00 \copyright2019 IEEE \hfill}
  \gdef\mycopyrightnotice{}
}

\title{Agglomerative Clustering of Handwritten Numerals to Determine Similarity of Different Languages}

\conf{2019 22nd International Conference on Computer and Information Technology (ICCIT), 18-20 December, 2019}

\author{\IEEEauthorblockN{Md. Rahat-uz-Zaman}
\IEEEauthorblockA{\textit{Dept. of Computer Science and Engineering} \\
\textit{Khulna University of Engineering \& Technology}\\
Khulna, Bangladesh \\
rahatzamancse@gmail.com}
\and
\IEEEauthorblockN{Shadmaan Hye}
\IEEEauthorblockA{\textit{Dept. of Computer Science and Engineering} \\
\textit{Khulna University of Engineering \& Technology}\\
Khulna, Bangladesh \\
praptishadmaan@gmaill.com}
}

\newcommand{\samplesize}{5.5mm}
\newcommand{\specialcell}[2][c]{
  \begin{tabular}[#1]{@{}c@{}}#2\end{tabular}}
\newcommand{\datasettable}{
\begin{table*}[!ht]
\caption{Datasets used for the proposed methodology.}
\centering
\resizebox{\textwidth}{!}{%
\begin{tabular}{|l|l|l|c|l|}
\hline
\textbf{S/L} & \multicolumn{1}{c|}{\textbf{Dataset}} & \multicolumn{1}{c|}{\textbf{Language}} & \textbf{Samples Used} & \multicolumn{1}{c|}{\textbf{Sample Images}} \\ \hline

1.a & \specialcell{ISI Bangla Numeral\\Dataset \cite{bhattacharya2008handwritten, bhattacharya2005databases}} & Bangla & 23,392 & \includegraphics[width=\samplesize, height=\samplesize]{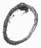}\includegraphics[width=\samplesize, height=\samplesize]{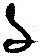}\includegraphics[width=\samplesize, height=\samplesize]{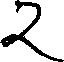}\includegraphics[width=\samplesize, height=\samplesize]{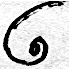}\includegraphics[width=\samplesize, height=\samplesize]{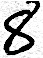}\includegraphics[width=\samplesize, height=\samplesize]{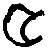}\includegraphics[width=\samplesize, height=\samplesize]{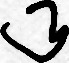}\includegraphics[width=\samplesize, height=\samplesize]{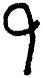}\includegraphics[width=\samplesize, height=\samplesize]{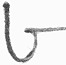}\includegraphics[width=\samplesize, height=\samplesize]{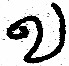} \\ \hline

1.b & NumtaDB \cite{alam2018numtadb} & Bangla & 85,000+ & \includegraphics[width=\samplesize, height=\samplesize]{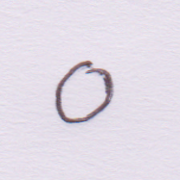}\includegraphics[width=\samplesize, height=\samplesize]{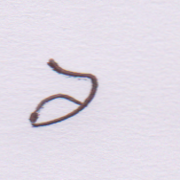}\includegraphics[width=\samplesize, height=\samplesize]{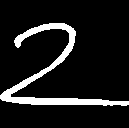}\includegraphics[width=\samplesize, height=\samplesize]{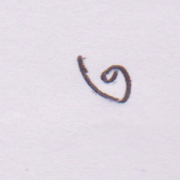}\includegraphics[width=\samplesize, height=\samplesize]{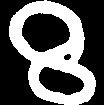}\includegraphics[width=\samplesize, height=\samplesize]{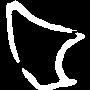}\includegraphics[width=\samplesize, height=\samplesize]{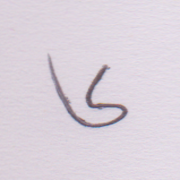}\includegraphics[width=\samplesize, height=\samplesize]{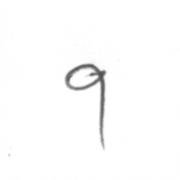}\includegraphics[width=\samplesize, height=\samplesize]{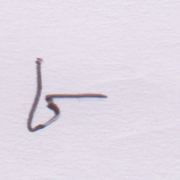}\includegraphics[width=\samplesize, height=\samplesize]{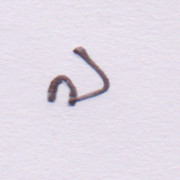} \\ \hline

2 & MNIST \cite{deng2012mnist} & English & 60,000 & \includegraphics[width=\samplesize, height=\samplesize]{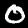}\includegraphics[width=\samplesize, height=\samplesize]{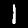}\includegraphics[width=\samplesize, height=\samplesize]{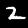}\includegraphics[width=\samplesize, height=\samplesize]{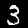}\includegraphics[width=\samplesize, height=\samplesize]{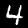}\includegraphics[width=\samplesize, height=\samplesize]{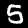}\includegraphics[width=\samplesize, height=\samplesize]{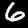}\includegraphics[width=\samplesize, height=\samplesize]{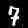}\includegraphics[width=\samplesize, height=\samplesize]{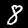}\includegraphics[width=\samplesize, height=\samplesize]{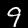} \\ \hline

3 & Chinese-Mandarin & Chinese & 100 & \includegraphics[width=\samplesize, height=\samplesize]{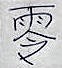}\includegraphics[width=\samplesize, height=\samplesize]{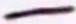}\includegraphics[width=\samplesize, height=\samplesize]{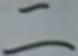}\includegraphics[width=\samplesize, height=\samplesize]{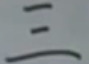}\includegraphics[width=\samplesize, height=\samplesize]{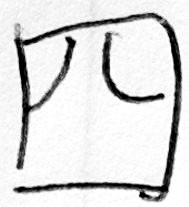}\includegraphics[width=\samplesize, height=\samplesize]{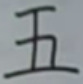}\includegraphics[width=\samplesize, height=\samplesize]{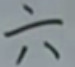}\includegraphics[width=\samplesize, height=\samplesize]{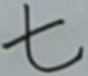}\includegraphics[width=\samplesize, height=\samplesize]{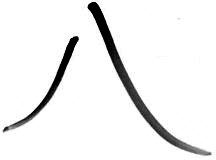}\includegraphics[width=\samplesize, height=\samplesize]{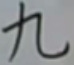} \\ \hline

4 & 278 Gujarati responses & Gujarati & 278 & \includegraphics[width=\samplesize, height=\samplesize]{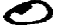}\includegraphics[width=\samplesize, height=\samplesize]{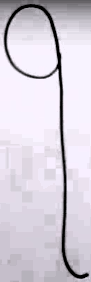}\includegraphics[width=\samplesize, height=\samplesize]{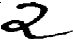}\includegraphics[width=\samplesize, height=\samplesize]{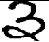}\includegraphics[width=\samplesize, height=\samplesize]{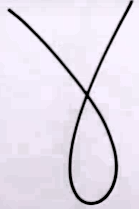}\includegraphics[width=\samplesize, height=\samplesize]{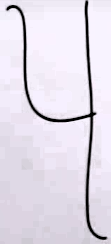}\includegraphics[width=\samplesize, height=\samplesize]{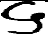}\includegraphics[width=\samplesize, height=\samplesize]{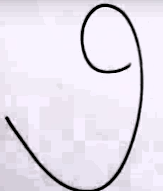}\includegraphics[width=\samplesize, height=\samplesize]{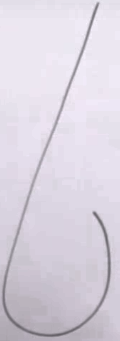}\includegraphics[width=\samplesize, height=\samplesize]{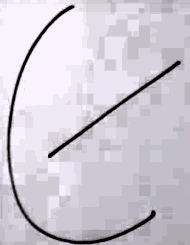} \\ \hline

5 & CMATERdb 3.3.1 \cite{das2012statistical} & Arabic & 3,000 & \includegraphics[width=\samplesize, height=\samplesize]{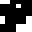}\includegraphics[width=\samplesize, height=\samplesize]{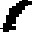}\includegraphics[width=\samplesize, height=\samplesize]{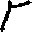}\includegraphics[width=\samplesize, height=\samplesize]{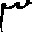}\includegraphics[width=\samplesize, height=\samplesize]{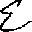}\includegraphics[width=\samplesize, height=\samplesize]{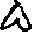}\includegraphics[width=\samplesize, height=\samplesize]{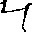}\includegraphics[width=\samplesize, height=\samplesize]{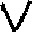}\includegraphics[width=\samplesize, height=\samplesize]{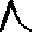}\includegraphics[width=\samplesize, height=\samplesize]{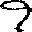} \\ \hline

6.a & CMATERdb 3.2.1 \cite{das2012statistical} & Devanagari & 150 & \includegraphics[width=\samplesize, height=\samplesize]{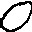}\includegraphics[width=\samplesize, height=\samplesize]{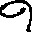}\includegraphics[width=\samplesize, height=\samplesize]{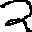}\includegraphics[width=\samplesize, height=\samplesize]{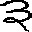}\includegraphics[width=\samplesize, height=\samplesize]{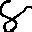}\includegraphics[width=\samplesize, height=\samplesize]{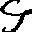}\includegraphics[width=\samplesize, height=\samplesize]{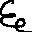}\includegraphics[width=\samplesize, height=\samplesize]{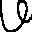}\includegraphics[width=\samplesize, height=\samplesize]{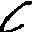}\includegraphics[width=\samplesize, height=\samplesize]{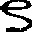} \\ \hline
 
6.b & \specialcell{ISI Devanagari Numeral\\Dataset \cite{bhattacharya2008handwritten, bhattacharya2005databases}} & Devanagari & 22,556 & \includegraphics[width=\samplesize, height=\samplesize]{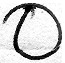}\includegraphics[width=\samplesize, height=\samplesize]{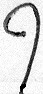}\includegraphics[width=\samplesize, height=\samplesize]{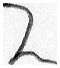}\includegraphics[width=\samplesize, height=\samplesize]{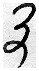}\includegraphics[width=\samplesize, height=\samplesize]{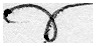}\includegraphics[width=\samplesize, height=\samplesize]{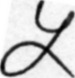}\includegraphics[width=\samplesize, height=\samplesize]{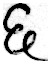}\includegraphics[width=\samplesize, height=\samplesize]{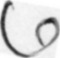}\includegraphics[width=\samplesize, height=\samplesize]{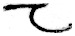}\includegraphics[width=\samplesize, height=\samplesize]{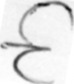} \\ \hline

7 & Thai web collection & Thai & 100 & \includegraphics[width=\samplesize, height=\samplesize]{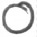}\includegraphics[width=\samplesize, height=\samplesize]{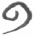}\includegraphics[width=\samplesize, height=\samplesize]{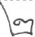}\includegraphics[width=\samplesize, height=\samplesize]{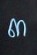}\includegraphics[width=\samplesize, height=\samplesize]{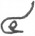}\includegraphics[width=\samplesize, height=\samplesize]{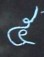}\includegraphics[width=\samplesize, height=\samplesize]{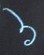}\includegraphics[width=\samplesize, height=\samplesize]{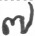}\includegraphics[width=\samplesize, height=\samplesize]{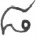}\includegraphics[width=\samplesize, height=\samplesize]{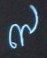} \\ \hline

8 & Burmese 278 responses & Burmese & 278 & \includegraphics[width=\samplesize, height=\samplesize]{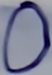}\includegraphics[width=\samplesize, height=\samplesize]{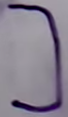}\includegraphics[width=\samplesize, height=\samplesize]{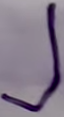}\includegraphics[width=\samplesize, height=\samplesize]{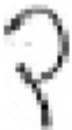}\includegraphics[width=\samplesize, height=\samplesize]{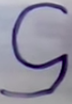}\includegraphics[width=\samplesize, height=\samplesize]{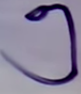}\includegraphics[width=\samplesize, height=\samplesize]{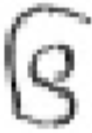}\includegraphics[width=\samplesize, height=\samplesize]{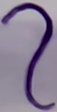}\includegraphics[width=\samplesize, height=\samplesize]{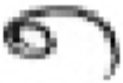}\includegraphics[width=\samplesize, height=\samplesize]{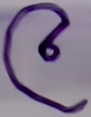} \\ \hline

9 & Kannada-MNIST \cite{prabhu2019kannada} & Kannada & 60,000 & \includegraphics[width=\samplesize, height=\samplesize]{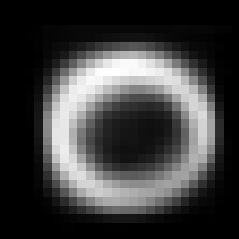}\includegraphics[width=\samplesize, height=\samplesize]{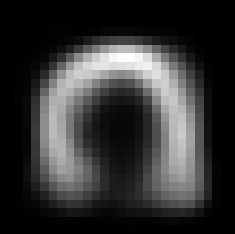}\includegraphics[width=\samplesize, height=\samplesize]{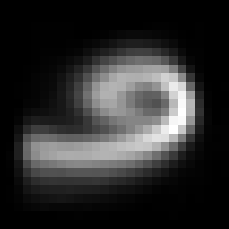}\includegraphics[width=\samplesize, height=\samplesize]{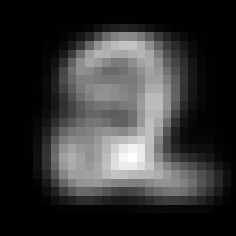}\includegraphics[width=\samplesize, height=\samplesize]{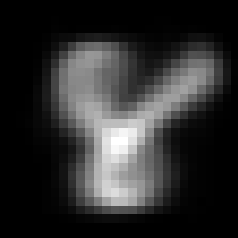}\includegraphics[width=\samplesize, height=\samplesize]{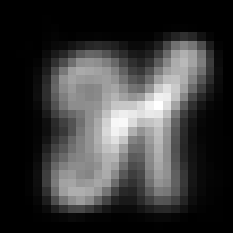}\includegraphics[width=\samplesize, height=\samplesize]{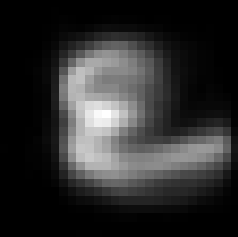}\includegraphics[width=\samplesize, height=\samplesize]{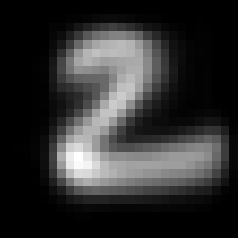}\includegraphics[width=\samplesize, height=\samplesize]{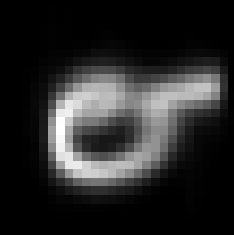}\includegraphics[width=\samplesize, height=\samplesize]{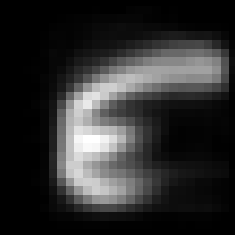} \\ \hline

10 & Japanese-Kanji & Japanese & 100 & \includegraphics[width=\samplesize, height=\samplesize]{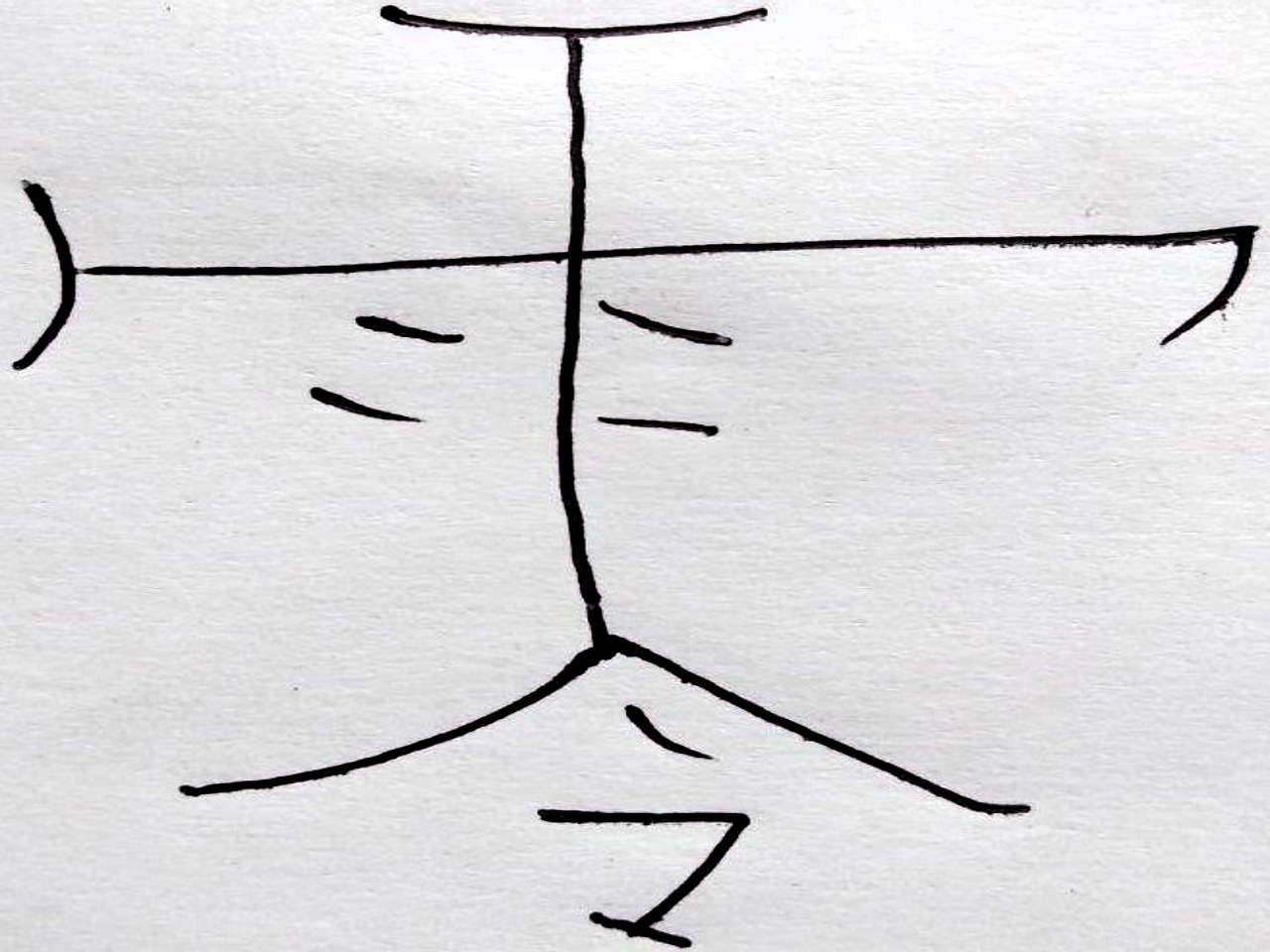}\includegraphics[width=\samplesize, height=\samplesize]{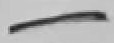}\includegraphics[width=\samplesize, height=\samplesize]{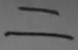}\includegraphics[width=\samplesize, height=\samplesize]{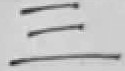}\includegraphics[width=\samplesize, height=\samplesize]{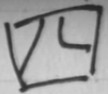}\includegraphics[width=\samplesize, height=\samplesize]{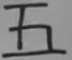}\includegraphics[width=\samplesize, height=\samplesize]{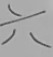}\includegraphics[width=\samplesize, height=\samplesize]{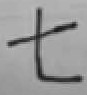}\includegraphics[width=\samplesize, height=\samplesize]{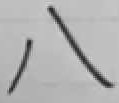}\includegraphics[width=\samplesize, height=\samplesize]{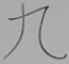} \\ \hline

11 & \specialcell{CMATERdb 3.4.1\\ \cite{das2012genetic, das2012statistical, das2012novel, das2009handwritten, das2015improved, das2014benchmark}} & Telugu & 4,000 & \includegraphics[width=\samplesize, height=\samplesize]{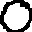}\includegraphics[width=\samplesize, height=\samplesize]{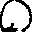}\includegraphics[width=\samplesize, height=\samplesize]{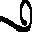}\includegraphics[width=\samplesize, height=\samplesize]{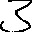}\includegraphics[width=\samplesize, height=\samplesize]{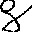}\includegraphics[width=\samplesize, height=\samplesize]{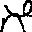}\includegraphics[width=\samplesize, height=\samplesize]{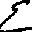}\includegraphics[width=\samplesize, height=\samplesize]{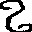}\includegraphics[width=\samplesize, height=\samplesize]{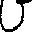}\includegraphics[width=\samplesize, height=\samplesize]{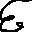} \\ \hline

12 & Tamil web collection & Tamil & 100 & \includegraphics[width=\samplesize, height=\samplesize]{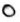}\includegraphics[width=\samplesize, height=\samplesize]{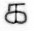}\includegraphics[width=\samplesize, height=\samplesize]{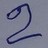}\includegraphics[width=\samplesize, height=\samplesize]{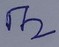}\includegraphics[width=\samplesize, height=\samplesize]{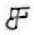}\includegraphics[width=\samplesize, height=\samplesize]{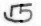}\includegraphics[width=\samplesize, height=\samplesize]{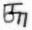}\includegraphics[width=\samplesize, height=\samplesize]{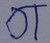}\includegraphics[width=\samplesize, height=\samplesize]{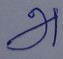}\includegraphics[width=\samplesize, height=\samplesize]{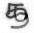} \\ \hline

13 & Malayalam-db & Malayalam & 100 & \includegraphics[width=\samplesize, height=\samplesize]{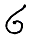}\includegraphics[width=\samplesize, height=\samplesize]{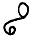}\includegraphics[width=\samplesize, height=\samplesize]{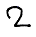}\includegraphics[width=\samplesize, height=\samplesize]{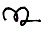}\includegraphics[width=\samplesize, height=\samplesize]{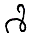}\includegraphics[width=\samplesize, height=\samplesize]{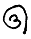}\includegraphics[width=\samplesize, height=\samplesize]{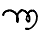}\includegraphics[width=\samplesize, height=\samplesize]{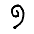}\includegraphics[width=\samplesize, height=\samplesize]{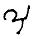}\includegraphics[width=\samplesize, height=\samplesize]{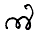} \\ \hline

\end{tabular}
}
\label{tab:datasets}
\end{table*}
}

\begin{document}

\maketitle

\begin{abstract}
Handwritten numerals of different languages have various characteristics. Similarities and dissimilarities of the languages can be measured by analyzing the extracted features of the numerals. Handwritten numeral datasets are available and accessible for many renowned languages of different regions. In this paper, several handwritten numeral datasets of different languages are collected. Then they are used to find the similarity among those written languages through determining and comparing the similitude of each handwritten numerals. This will help to find which languages have the same or adjacent parent language. Firstly, a similarity measure of two numeral images is constructed with a Siamese network. Secondly, the similarity of the numeral datasets is determined with the help of the Siamese network and a new random sample with replacement similarity averaging technique. Finally, an agglomerative clustering is done based on the similarities of each dataset. This clustering technique shows some very interesting properties of the datasets. The property focused in this paper is the regional resemblance of the datasets. By analyzing the clusters, it becomes easy to identify which languages are originated from similar regions.
\end{abstract}


\section{Introduction}
The evolution of writing has grown from tokens to pictography, syllabary and then alphabets. These writings have started independently in Near East, China and Mesoamerica. In different regions, different writing techniques have been developed. In the current stage, it is very difficult to analyze which handwritten numerals have been originated from similar regions.
\par
Due to the fact of increasing tasks of handwritten numeral recognition analysis, formation of several datasets regarding this can be found for almost all the languages. This paper has tried to collect, create, organize and use these datasets to find the similarity of the writing patterns of the handwritten numerals of one dataset of a language, to the handwritten numerals of another dataset of another language. Then these similarities from one dataset to another have been resulted in an agglomerative clustering of datasets, which has revealed the visual resemblance of one dataset to another.
\par
This whole technique does not only measure the similarity of the size and shape of each character but also takes into account other features such as the writing techniques. Writing techniques in each written digit includes where the stroke width is thin or where it is thick. This is because deep learning is used which automatically extracts as many features as possible. 
\par
The clustering of these datasets can be used to identify the origins of a particular language, or to be precise the numerals, knowing the origin of the languages of other datasets close to it. For example, if one knows the region where language `A' is used, then he can know the region of languages `B', `C', etc if they are close to `A'.
\par
The proposed methodology in this paper is divided into 2 parts. The first part is named as random sample similarity averaging to get similarity of two datasets discussed in section \ref{sec:similarity} with the help of Siamese network which is created and proposed on section \ref{sec:siamese}. The second part is the clustering of the datasets using the similarity results of the first part which is discussed in section \ref{sec:cluster}.

\section{Literature Review}
Similar work of this paper, which detect similarity of handwriting or language could not be found anywhere. So, some works the authors thought to be close is mentioned here for benchmarking the proposed methodology. Bin Zhang et al. \cite{binaryvectordis2003} has tried to choose a dissimilarity measure on different handwriting identifications. They have characterized the identifications with the help of binary micro-features and were successful to pertain 8 dissimilarity measures.
\par
Zhuoyao Zhong \cite{zhong2016spottingnet} has created a neural network named SpottingNet which is used to detect similarities between word images. They have used George Washington (GW) dataset \cite{rath2003word} and have achieved 80.03\% mAP.
\par
Ghada Sokar \cite{sokar2018generic} has used support vector machine (SVM) and Siamese network one-shot classification for optical character recognition (OCR) and gained almost similar result to state of the art CNN model.
\par
In this paper, similarity is detected between language numerals (not between just two images) which is different from the works mentioned previously. In  order to be general, multiple datasets of the same language are also taken.
\par
The methodology, along with sample datasets used, experiments and results are provided in the rest of the paper. The next section discusses the proposed Siamese network architecture. Then, after discussing the two major parts of the methodology, the following sections demonstrate how the result for the selected datasets have been attained. Then the paper is concluded with some applications, critics and discussion about the work in this paper.

\section{Siamese Network\label{sec:siamese}}
The working system is designed having a primary assumption. Two origins of two written languages are close if the datasets of those languages have high similarity between themselves. Usually, if two languages derive from a same older language, then the process of derivation is mostly a slight deformation of the parent language. Therefore, if two languages are very similar to each other, there is a high probability that they are slightly deformed from a same parent language. To find the similarity of two languages, first it is needed to find similarity between two numerals or images. This requires a similarity function or a measuring technique.
\par
\begin{figure}
\centerline{\includegraphics[width=0.8\linewidth]{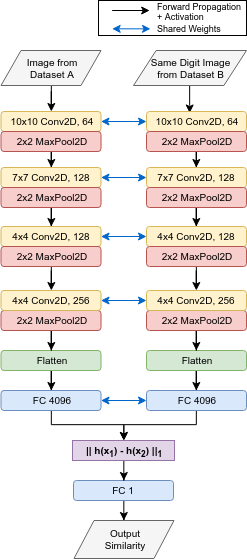}}
\caption{Architecture of Siamese Network.}
\label{fig:siamese}
\end{figure}

This similarity function used in this paper is a Siamese network \cite{bromley1994signature} from the proposed architecture of Gregory Koch \cite{koch2015siamese}. The architecture of the Siamese network is presented on figure \ref{fig:siamese}. Siamese network is a network which uses two instances of a same neural network (hence called Siamese or `twin') to extract a one dimensional feature vector from an image of 1 (gray-scale) or 3 channels (RGB) and then measuring the distance between the feature vectors. Convolutional Layers \cite{lecun1995convolutional} tend to perform significantly better result on image data, so they are used in Siamese network. For each of these layers, rectified linear unit (ReLU) \cite{xu2015empirical} is used as the activation function. Sigmoid \cite{han1995influence} is used for the last two dense layers to make the output between 0 and 1. The loss function used in this experiment is Binary Cross Entropy loss function. Adam \cite{kingma2014adam} optimizer is used as the optimizer of the loss function. The image input size was fixed to 105x105 gray-scale image. The training of the network is done on only one dataset of the selected numeral datasets.

\section{Random sampled similarity averaging of two datasets\label{sec:similarity}}
Siamese network will only work for two images, not for two datasets. In this part of the methodology, the technique of how the similarity of two datasets can be measured is discussed. Each dataset has a lot of images each labeled as a digit from 0 to 9. A tuple of two images is created where each image comes from the two datasets of whose similarity is needed to measure. Then the similarity of these two images is determined with the previously established similarity measuring technique in section \ref{sec:siamese}. 
\par
\begin{figure}
\centerline{\includegraphics[width=\linewidth, height=5.58in]{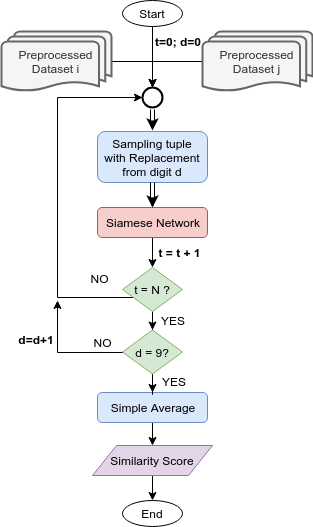}}
\caption{Flowchart of determining similarity of two datasets.}
\label{fig:similarity}
\end{figure}

Establishing the similarity measure of two datasets depends on some more predefined parameters. The input image size is an important parameter which is selected to be 105x105. A sample size is fixed for the number of comparisons of each digit which will be done with the Siamese network. Suppose the sample size is N. For each digit, N tuples are sampled with replacement to compare and obtain the similarity with the network. Then all these $10*N$ values will be averaged to get the final similarity score between the two datasets. This process is iteratively continued for all combinations of the datasets taken into account. If number of datasets used is $M$, then total number of iterations needed to compute the similarity matrix is $\binom{M}{2}$. The whole process of measuring the similarity is demonstrated on figure \ref{fig:similarity} as a flowchart.

\section{Clustering of datasets\label{sec:cluster}}
From section \ref{sec:similarity}, a symmetric matrix of similarity of the used datasets is achieved. As no feature vector is generated from any dataset, the clustering is needed to be done on the basis of relative position from each other. Instead of distance, similarity measure is used for the clustering. For all these reasons, a dendrogram is used to represent the clustering. The dendrogram of the clustering is presented in figure \ref{fig:dendrogram}.
\par
Unweighted average linkage clustering (UPGMA) \cite{sokal1958statistical} is used as the linkage criterion for the clustering. This determines the similarity between clusters as an averaging function of the pairwise similarity. The similarity measuring equation is provided in \ref{eq:linkage}. Here function $\phi$ represents the similarity of two datasets $A$ and $B$.

\begin{equation}
\phi(A, B)=\frac{1}{|A||B|} \sum_{a \in A} \sum_{b \in B} S(a, b)
\label{eq:linkage}
\end{equation}

The distance function was first chosen by inverting the similarity function. But it is observed that using the similarity function without converting it to distance function provides better result.

\section{Experiments and Results}
\subsection{Datasets Used}
All the datasets used in this paper are presented briefly in table \ref{tab:datasets}. As not all numeral datasets of each language is publicly available, some datasets were created for this research purpose. Siamese network is a one-shot learner, as a result smaller number of images makes negligible degradation in the expected result.


Some large datasets are created by CVPR unit, Indian Statistical Institute, Kolkata on the language Bangla and Devanagari. ISI Bangla dataset contains total 23,392 gray-scale images of Bangla handwritten numerals and ISI Devanagari dataset contains 22,556 gray-scale images. All the images are handwritten with black ink on a white background. These are gathered from postal mail pieces' pin codes.
\par
A comparatively lower quality of datasets are created by Computer Science and Engineering Department, Jadavpur University, Kolkata named Center for Microprocessor Application for Training Education and Research database (CMATERdb). The size of the images are all 32x32 and in bmp format. This dataset is used for Arabic, Devanagari and Telugu languages.
\par
The authors could not find any well established dataset for Chinese, Japanese, Thai, Tamil and Malayalam numerals. Hence, they have created their own dataset by writing and collecting images from the internet. 100 images were collected for each of the datasets since not much images are needed for the methodology to work. There are a lot of variants in both Chinese and Japanese languages. For the experiment, Mandarin Chinese and Japanese Kanji are used.

\datasettable

\subsection{Preprocessing of Datasets}
\begin{figure}[!b]
\centerline{\includegraphics[width=\linewidth]{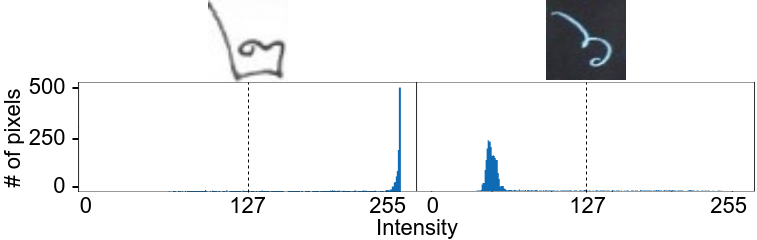}}
\caption{Image background generalization with histogram observation.}
\label{fig:histogram}
\end{figure}

Because the preprocessing part is dependent solely on the characteristics of the images in the datasets used, this subsection is included in the experiment's section. Firstly, some datasets (NumtaDB, Thai web collection) have a mixture of both white numeral on black background, and black numeral on white background. Because of that reason the images are classified by inspecting the histogram of the images. A sample histogram is provided in \ref{fig:histogram}. It can be observed from histograms of selected images that, the background spans more area in the image than the foreground because the foreground is only the pen stroke taking a small number of pixels. So the classification is done by \ref{eq:histogram}.

\begin{equation}
\text{bg}(\text{img}) = 
    \begin{cases}
    \text{white if} \sum\limits_{i = 0}^{127} \text{hist}(img, i) \leq \sum\limits_{i = 128}^{255} \text{hist}(img, i) \\
    \text{black if} \sum\limits_{i = 0}^{127} \text{hist}(img, i) > \sum\limits_{i = 128}^{255} \text{hist}(img, i) \\
    \end{cases}
\label{eq:histogram}
\end{equation}

The image is transformed using equation \ref{eq:histogram} such that, the numerals are written in white colour having a black background. OTSU thresholding is applied to make the image binary. This will make the Siamese networks work easier as it has to  take input of 2 intensity values (0 and 1). The images are of different sizes in different datasets. The final preprocessing step would be to resize the images to a fixed input size. We have taken the input size to be of 105 pixels width and 105 pixels height. Bilinear interpolation is used to upsample the images.

\subsection{Implementation Details}
The whole programming is done with the python programming language and bash shell scripting. Keras \cite{chollet2015keras} and PyTorch \cite{paszke2017automatic} is used as the deep learning framework. Plotly \cite{plotly} and Matplotlib \cite{Hunter:2007} is used to visualise the figures in this paper. For the preprocessing and image resizing, OpenCV \cite{opencv_library} is used. Numpy \cite{numpy} python package is used to do all the mathematical operations such as averaging and histogram regional summation. The hierarchical clustering is done with Sci-kit learn \cite{Pedregosa:2011:SML:1953048.2078195} python framework. The network model is trained on a custom built PC which has CPU Intel Core i7 with 3.5 GHz speed and RAM 32 GB with GPU NVIDIA GeForce GTX 1080 Ti in a Jupyter Notebook environment.

\begin{figure*}
\centerline{\includegraphics[width=0.8\linewidth]{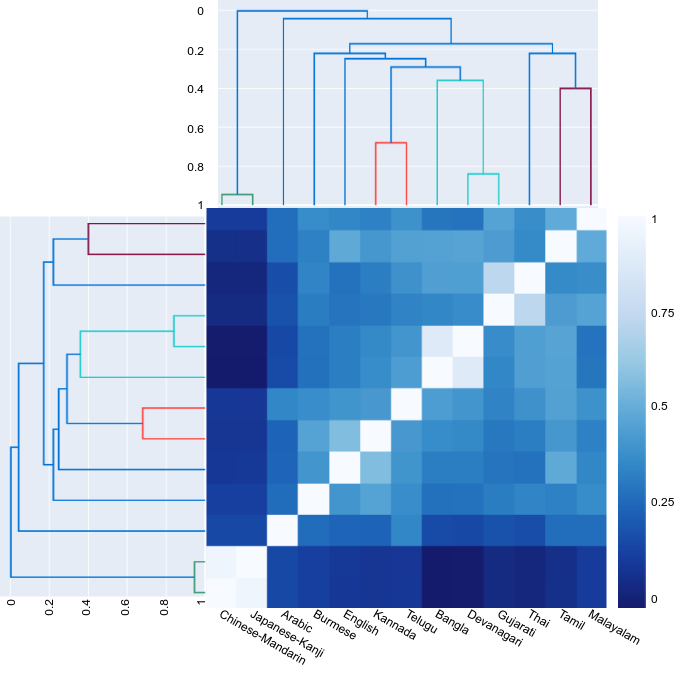}}
\caption{Dendrogram of clusters of the datasets.}
\label{fig:dendrogram}
\end{figure*}

\subsection{Result Analysis\label{sec:result}}
The agglomerative clustering creates a bunch of clusters of different languages based on the similarity of each other. The resulting similarity matrix is graphically presented on figure \ref{fig:dendrogram} where complete white represents 1 (almost identical) and dark blue represents 0 (not similar at all). Based on the matrix, the clustering is also done and the dendrogram is presented in the same figure \ref{fig:dendrogram}.

\begin{figure}[!h]
\centerline{\includegraphics[width=\linewidth, height=1.5in]{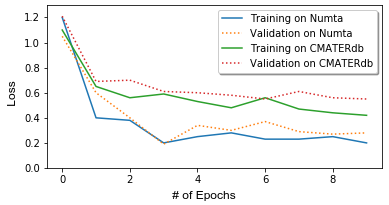}}
\caption{Epoch vs Loss plot in training the Siamese network on two datasets separately.}
\label{fig:train}
\end{figure}

Training the Siamese network was a very important task. Generally, Siamese networks are trained until the accuracy is reached at maximum \cite{NIPS2013_5166} without overfitting the model. But, training on a single dataset will not work for other datasets if the network is trained too much on that dataset alone. So, the authors have decided to train the network slightly less and get a low accuracy instead. This will keep the model workable on measuring similarity between different datasets. As training requires a lot of images, only few datasets are eligible for the training. NumtaDB is taken to train the Siamese network. Figure \ref{fig:train} shows the loss of Siamese network while training the network with NumtaDB and CMATERdb 3.4.1.

Analyzing the dendrogram on figure \ref{fig:dendrogram} shows that the most similar languages are Japanese Kanji and Mandarin Chinese. In fact, if the numerals are viewed one by one, it would be hard for anyone to differentiate between the two numerals of Chinese and Japanese. Then the most similar languages are Gujarati and Devanagari. This is because Gujarati language is originated from Devanagari. On the other hand, Kannada and Telugu are also very similar. The regions using these languages are very close to each other. Hence, the clustering appears to be working as assumed in the first place. Almost all the languages which are originated in Asia are observed to be merged earlier with the closer adjacent languages than other languages in the dendrogram.

\section{Discussion and Conclusion}
This paper has tried to identify which languages are similar to each other in different aspects. In spite of being very realistic, the resulting cluster in section \ref{sec:result} has some flaws. English has merged with other Asian languages before Burmese where Burmese, Devanagari and Gujarati are more similar than English. Mandarin Chinese are Japanese Kanji seem to be very close to each other in the dendrogram. But only the numerals of the languages are same. Words are different from each other in the two languages.
\par
This paper has tried to use only 13 different sample languages so far. For future work, more languages are considered to be included. The main focus was only on the Asian languages. English was the only language taken from continent Europe because of the global popularity and ease of collection of the dataset. Moreover, the network needs to be more tuned to fit the datasets with even more accuracy.

\section{Acknowledgement}
The authors want to first express their profound gratitude and heartiest thanks to all who have inspired them for the contribution of this paper. Authors also want to thank all the people who have helped them to find and collect images of handwritten numerals of several languages. Lastly, the authors want to thank the reviewers of this paper and to the researchers on the field of machine learning for contributing to the world of knowledge.

\bibliography{Reference.bib}

\end{document}